# Predicting Heart Disease and Reducing Survey Time Using Machine Learning Algorithms

Salahaldeen Rababa[1], Asma Yamin[1], Shuxia Lu[1] and Ashraf Obaidat[1]

[1]State University of New York at Binghamton, Binghamton, New York,  srababa1@binghamton.edu, Ayamin1@binghamton.edu, slu@binghamton.edu, aobeida3@binghamton.edu

Corresponding author's Email: srababa1@binghamton.edu

**Abstract:** Currently, many researchers and analysts are working toward medical diagnosis enhancement for various diseases. Heart disease is one of the common diseases that can be considered a significant cause of mortality worldwide. Early detection of heart disease significantly helps in reducing the risk of heart failure. Consequently, the Centers for Disease Control and Prevention (CDC) conducts a health-related telephone survey yearly from over 400,000 participants. However, several concerns arise regarding the reliability of the data in predicting heart disease and whether all of the survey questions are strongly related. This study aims to utilize several machine learning techniques, such as support vector machines and logistic regression, to investigate the accuracy of the CDC's heart disease survey in the United States. Furthermore, we use various feature selection methods to identify the most relevant subset of questions that can be utilized to forecast heart conditions. To reach a robust conclusion, we perform stability analysis by randomly sampling the data 300 times. The experimental results show that the survey data can be useful up to 80% in terms of predicting heart disease, which significantly improves the diagnostic process before bloodwork and tests. In addition, the amount of time spent conducting the survey can be reduced by 77% while maintaining the same level of performance.

## 1. Introduction

Heart Disease (HD) is a major public health concern [1]. According to the World Health Organization (WHO), one-third of global deaths are from heart and cardiac disease [2]. High cholesterol, obesity, high triglycerides, hypertension, and other unhealthy behaviors increase heart disease risk. The presence of heart disease can reduce blood flow and cause heart failure [3]. The high death rate is due to the difficulty of diagnosing heart disease, especially without advanced technology and medical specialists [4]. Heart disease requires blood pressure, ECG, auscultation, cholesterol, and blood sugar tests. These tests take a long time, delaying the patient's medication and harming them. Researchers used data mining and machine learning (ML) algorithms to develop heart disease prediction systems. Current methods for detecting heart disease are ineffective due to inaccuracy and execution time [5].

Machine Learning (ML) prediction systems start with a dataset. Medical facilities and organizations have begun collecting patient data to improve service quality, but the data is insufficient and noisy. Data collection, especially survey data, is time-consuming. Researchers used medical data to predict diseases. Others built prediction models from survey data. Others built prediction models from survey data [6,7]. The Behavioral Risk Factor Surveillance System (BRFSS) surveys over 400,000 participants annually [8]. This study aims to shorten surveys by selecting the most important questions while maintaining performance. We examine the survey's ability to predict heart disease. We build several prediction models and evaluate their performance before and after finding the most relevant set of features.

## 2. Related Work

Machine learning algorithms have been proposed in several research papers by researchers to predict the heart diseases in early stages, using different machine learning techniques. Mohan et. al. 2019 applied machine learning technique for the prediction improvement of cardiovascular disease. The machine learning techniques used in the paper were K-Nearest Neighbor Algorithm (KNN), Decision Trees (DT), Genetic algorithm (GA), and Naïve Bayes (NB). The results with an accuracy more than 88% was reached on a dataset retrieved from heart disease classification of the Cleveland UCI repository [9]. Li et al. 2020 proposed a system that uses machine learning techniques to diagnose heart disease. The classification algorithms used to develop the proposed system are Support Vector Machine (SVM), Logistic Regression (LR), Artificial Neural Network (ANN), KNN, NB, and DT [10]. Dua et al. 2019 applied feature selection algorithms. The feature selection algorithms that have been applied the Relief, Minimal redundancy maximal relevance, least absolute shrinkage selection operator and local learning [11].

Also, Li et al. 2020 proposed a new feature selection algorithm called fast conditional mutual information. The results of the developed system based on the selected features on a dataset retrieved from Cleveland Heart Disease were high compared to other used methods [12]. Khourdifi et al. 2019 proposed a hybrid method for heart disease classification. The proposed method included various machine learning techniques such as KNN, SVM, NB, RF and a Multilayer Perception Neural Network (MLNN) optimized by Particle Swarm Optimization (PSO) combined with Ant Colony Optimization (ACO) approaches. For the feature selection, the Fast Correlation-Based method was used. The results for the proposed system were compared using different performance measures and showed more than 99% accuracy [13]. All the related papers showed good results in heart disease prediction. However, this study aims to see how much we can rely on the survey as the first step of heart disease diagnosis before moving on to blood work and other tests using the common risk factors addressed in the dataset. We are also working on using different embedded feature selection and classification methods by building several prediction models and evaluating their performance before and after finding the most relevant set of features. In addition, we applied data sampling techniques to solve the imbalanced data problem. Finally, we investigated how much reduction in the survey time after using a subset question rather than complete survey questions while maintaining the model's performance measures.

## 3. Methodology

This section introduces the methodologies for building our prediction models using several machine learning algorithms, selecting the most stable list of questions through feature selection methods, and reducing survey time. Figure 1 shows the methodology steps to determine how much we can rely on the initial survey to diagnose heart disease before proceeding for further heart disease tests. Figure 2 depicts the methodology steps on how to select the most stable subset of questions to reduce survey time. In the next subsections, we present the implementation of our methodology in more details.

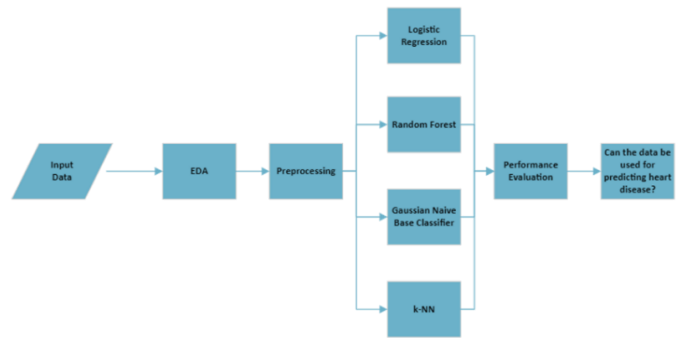

Figure 1. Prediction Models Methodology

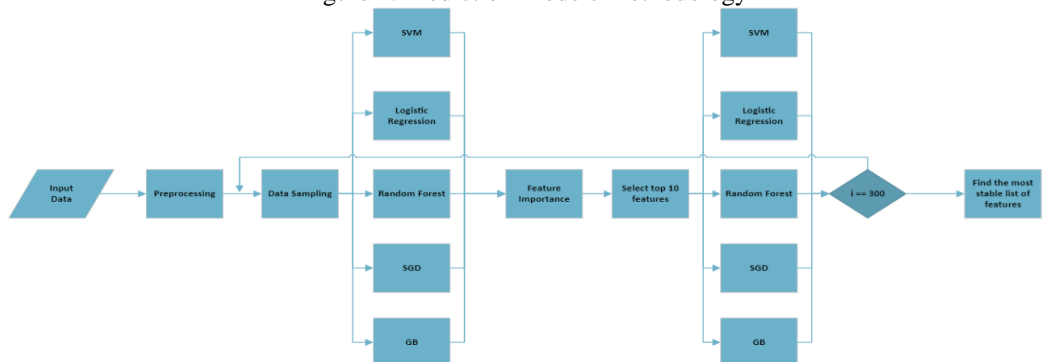

Figure 2. Feature Selection Methodology

### 3.1 Data preprocessing

Data preprocessing is an essential step in machine learning since the quality of the data and the information derived from it directly impact the performance measures of the suggested prediction model [14]. This step deals with outliers, continuous and categorical variables encoding, and normalization. Categorical variables are incompatible with most machine learning methods unless converted to numerical values. Also, many algorithms' results differ depending on how categorical variables are encoded. On the other hand, although continuous variables are easy to analyze, we need to consider the range of each one. Therefore, we used normalization to scale the numerical variables into the (0,1) range.

### 3.2 Data Sampling

Data sampling is a statistical analysis technique that involves selecting, manipulating, and analyzing a representative selection of data points to uncover patterns and trends in a sizeable imbalanced dataset [15]. It allows to build predictive models and deal with a limited, reasonable amount of data about a statistical population to swiftly create and run analytical models while still generating reliable results. In our approach, we randomly sampled the data such that each class can have the same number of observations.

### 3.3 Feature Selection and Prediction Models

After the data preparation step, machine learning algorithms can be utilized to perform the predication of the heart disease or attack. Since our aim is to classify the observations as whether they will experience a heart disease or attack, we delved more into supervised learning algorithms, classification algorithms in particular. We broadened our choice of algorithms to not only perform classification, but to include feature selection as well. Consequently, we used the following machine learning algorithms: k-Nearest Neighbor (KNN), Support Vector Machine (SVM), Logistic Regression (LR), Random Forest (RF), Stochastic Gradient Descent (SGD), and Gradient Boosting (GB) [16]. These algorithms also measure the feature importance in our data. That is, the importance of each question in the survey in terms of contributing to the class of interest, which in our case is heart disease or attack.

### 3.4 Stability Analysis

In some cases, the feature selection methods might produce different feature importance weights at each run. We refer to this problem as inconsistency of the results, where the first list of most important might be different from the second list of most important features for each corresponding run with the same input data and parameters [17]. To tackle this issue, we performed a stability analysis to find the most stable list of questions that contribute the most to our class of interest. In total, for each beforementioned prediction model, we repeated the pipeline of our analysis 300 times, with different samples of data at each iteration and the same input parameters. To aggregate the results at a model level, we considered the top repeated list of questions. Then, at the pipeline level, we considered the top repeated features across the different models.

### 3.5 Performance Measures

To evaluate the performance of each model, we used the following metrics: accuracy, precision, sensitivity, specificity, and F1 score [18]. Accuracy provides information about the model's power in terms of overall predictions. However, in case of imbalanced data, further performance metrics are needed. To accurately evaluate our models, we considered the precision, sensitivity, specificity, and F1 score, which further measures the classification power of the prediction models.

## 4. Results and Discussion

In this section, we present the experimental results of our prediction models. We first evaluated the performance of the models based on the whole dataset, then we considered data sampling and measured the performance before and after performing feature selection. To ensure that our selected list of questions is reliable, we repeated the experiment 300 times and recorded the top repeated features across all models as the most stable list of questions.

### 4.1 Data Description

We used the heart disease survey data published in Kaggle. The data included the results of conducting 253,680 telephone surveys by the Behavioral Risk Factor Surveillance System (BRFSS) [8]. These surveys are targeted at dealing with health conditions, hazards, and behaviors. There is a binary classification for heart disease in the data set, with 229,787 replies indicating that they do not have heart disease and 23,893 indicating that they do have heart disease. In addition, other 21 features/questions are recorded as well, such as BMI, diabetes, smoking, age, and income.

### 4.2 Classification Performance Based on all Data

Due to the imbalance in our data, the accuracy is not a good representative of the models' performance. Therefore, precision, sensitivity (recall), specificity, and F1 score are used, since they provide a clearer description of how well the prediction model can identify heart disease. Table 1 summarizes the performance of the models, for the training data set, RF had the highest recall of 99%, and for the testing data set LR is the best having 79% recall.

Table 1. Performance Measures of the Prediction Models Based on Whole Data Before and After Feature Selection

|     | Before feature selection | | | | After feature selection | | | |
| --- | --- | --- | --- | --- | --- | --- | --- | --- |
|     | Accuracy | Precision | Recall | F1 Score | Accuracy | Precision | Recall | F1 Score |
| LR  | 0.75 | 0.25 | 0.79 | 0.38 | 0.75 | 0.25 | 0.79 | 0.38 |
| KNN | 0.99 | 0.27 | 0.97 | 0.98 | 0.89 | 0.34 | 0.15 | 0.20 |
| GNB | 0.82 | 0.27 | 0.55 | 0.36 | 0.81 | 0.27 | 0.55 | 0.36 |
| RF  | 0.99 | 0.96 | 0.99 | 0.98 | 0.87 | 0.39 | 0.10 | 0.16 |

### 4.3 Data Sampling and feature selection

In this step, we applied data sampling by taking 1000 random samples from the negative dataset and 1000 random samples from the positive class dataset. After that, we applied LR, SVM, GB, RF, and SGB to extract the essential features; we repeated this process for 300 iterations. In each iteration, we computed how many times each feature was selected in each method. Figure 3 shows the percentage of each feature selected across all the iterations for each prediction model.

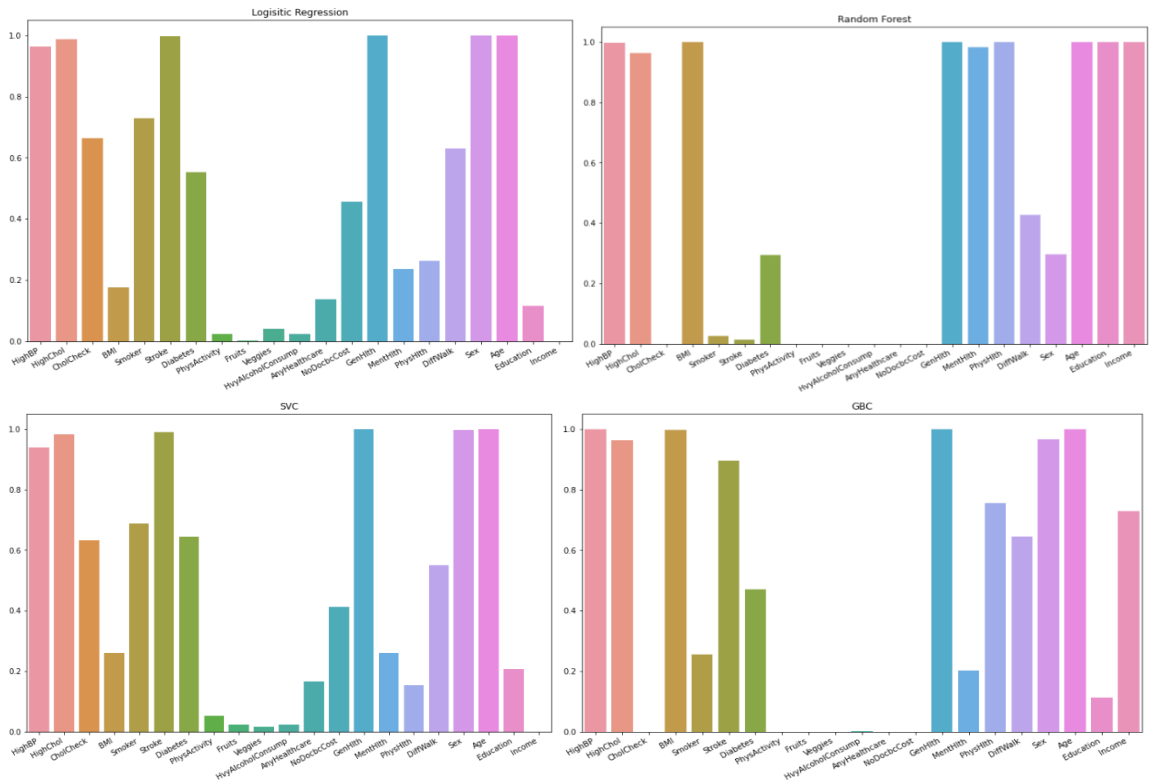

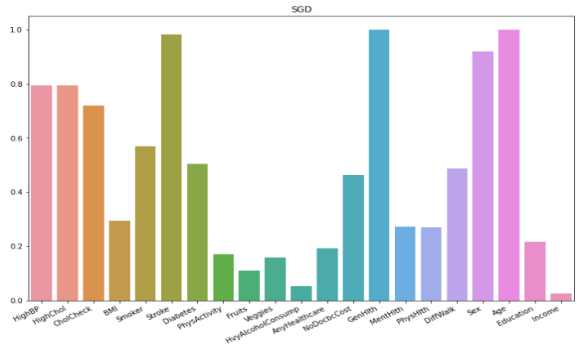

Figure 3. Feature Selection Methodology

### 4.4 Performance Measures After Feature Selection

Table 2 shows the performance measures for each prediction model. It is worth mentioning that after data sampling, the accuracy can be used as a representative performance measure since we use an equal sample size now for both testing and training datasets.

Table 2. Performance Measures of the Prediction Models Base on Data Sampling Before and After Feature Selection

|     | Before feature selection | | | | After feature selection | | | |
| --- | --- | --- | --- | --- | --- | --- | --- | --- |
|     | Accuracy | Precision | Recall | F1 Score | Accuracy | Precision | Recall | F1 Score |
| LR  | 0.77 | 0.76 | 0.79 | 0.78 | 0.76 | 0.73 | 0.78 | 0.76 |
| RF  | 1.0 | 1.0 | 1.0 | 1.0 | 0.74 | 0.69 | 0.82 | 0.75 |
| KNN | 0.79 | 0.78 | 0.82 | 0.80 | 0.77 | 0.75 | 0.80 | 0.77 |
| GNB | 0.84 | 0.82 | 0.88 | 0.85 | 0.75 | 0.72 | 0.81 | 0.76 |
| RF  | 0.76 | 0.77 | 0.76 | 0.76 | 0.74 | 0.73 | 0.73 | 0.73 |

The experimental results for the training dataset show that the RF prediction model has 100% accuracy among all models, and for the testing dataset, the SVM has 77% accuracy in identifying heart disease using half of the original features.

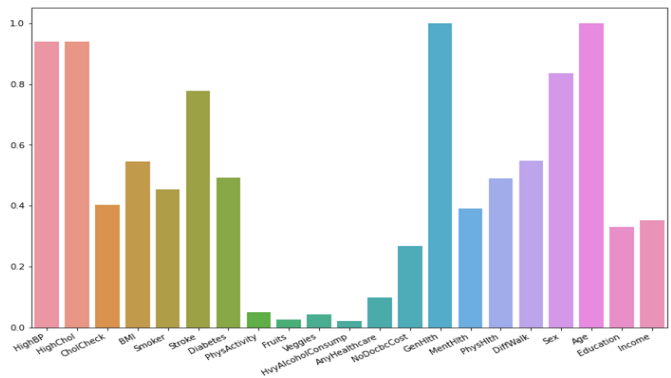

Figure 4. Feature Importance by all Classifiers

### 4.5 Reduction of Survey Time

To demonstrate the effect of selecting the top questions, we prepared two sets of survseys. The first one is similar to the original survey with (21) questions, and the second with only 10 questions, which are the top 10 repeated features/questions among all classifiers, shown in Figure 4. Those are questions related to General Health, Age, High Blood Pressure, High Cholesterol, Sex, Stroke, Difficulty Walking, BMI, Diabetes, and Physical Health. After selecting the most relevant questions during the feature selection process, we conducted a survey from two different groups of individuals; each group contained 15 individuals to avoid bias. The survey time was recorded for each person. After that, the conducted survey times were fitted using Expert fit software, and the goodness of fitness was measured using Akaike distance (AIC). Table 3 below shows that the survey time can be reduced by 77.9% after implementing feature selection.

Table 3. Survey Time Distribution Before and After Feature Selection

|  | Before feature selection | After feature selection |
|---|---|---|
| Number of Questions | 21 | 10 |
| Survey time distribution | Triangle (3.5,6.4,6.4) | Logistic (1.19,0.019) |
| Mean | 5.4 | 1.19 |
| Survey time reduction percent | - | 77.9% |

## 5. Conclusion

Machine learning techniques have been widely used in the field of data analytics in general, and in healthcare applications. The importance of machine learning cannot be overstated. In the context of healthcare applications, early and efficient heart disease diagnosis can help in early treatment for the patients. Using Machine learning techniques as a tool for predicting heart disease is considered one of the efficient ways of diagnosis. In this work, we applied several machine learning techniques to the widely available BRFSS dataset. Our aim was to test the reliability of the data in predicting heart disease and to search for the most stable list of questions that could be used to reduce the survey time. Our findings conclude that the data can be useful for up to 80% in predicting heart disease. Furthermore, selecting only the most important 10 questions can help reduce the survey time by 77% while maintaining comparable performance. In future work, we plan to extend our work by including other heart-related datasets. We also plan to consider other approaches for analysis and predictions, such as neural networks.